\def\year{2019}\relax
\documentclass[letterpaper]{article} 
\usepackage{aaai19}  
\usepackage{times}  
\usepackage{helvet}  
\usepackage{courier}  
\usepackage{url}  
\usepackage{graphicx}  
\usepackage{booktabs, multirow, multicol, bm}
\usepackage{soul}
\usepackage{tabularx}
\usepackage{latexsym}
\usepackage{subcaption}
\usepackage{amsmath,amsthm,amssymb}
\usepackage{algorithm,algorithmicx,algpseudocode}

\newcommand{\R}{\mathbb{R}}

\DeclareMathOperator*{\argmax}{arg\,max}

\newcommand{\citet}[1]
{\citeauthor{#1} ̃\shortcite{#1}}
\newcommand{\citep}{\cite}

\title{Learning Personalized End-to-End Goal-Oriented Dialog}
\author{
    Liangchen Luo$^\dagger$\thanks{This work was done when the first author was on an internship and the second author was a full-time employee researcher at Microsoft Research Asia.}, Wenhao Huang$^\ddagger$, Qi Zeng$^\dagger$, Zaiqing Nie$^\S$, Xu Sun$^\dagger$ \\
    $^\dagger$MOE Key Lab of Computational Linguistics, School of EECS, Peking University, Beijing, China \\
    $^\ddagger$Shanghai Discovering Investment, Shanghai, China $^\S$Alibaba AI Labs, Beijing, China \\
    $^\dagger${\tt \{luolc,pkuzengqi,xusun\}@pku.edu.cn} \\
    $^\ddagger${\tt huangwh@discoveringgroup.com} $^\S${\tt zaiqing.nzq@alibaba-inc.com}
}

\begin{document}
%

\maketitle
\begin{abstract}
    Most existing works on dialog systems only consider conversation content while neglecting the personality of the user the bot is interacting with, which begets several unsolved issues.
    In this paper, we present a personalized end-to-end model in an attempt to leverage personalization in goal-oriented dialogs.
    We first introduce a \textsc{Profile Model} which encodes user profiles into distributed embeddings and refers to conversation history from other similar users.
    Then a \textsc{Preference Model} captures user preferences over knowledge base entities to handle the ambiguity in user requests.
    The two models are combined into the \textsc{Personalized MemN2N}.
    Experiments show that the proposed model achieves qualitative performance improvements over state-of-the-art methods.
    As for human evaluation, it also outperforms other approaches in terms of task completion rate and user satisfaction.
\end{abstract}

\section{Introduction}
\label{sec:intro}
    
There has been growing research interest in training dialog systems with end-to-end models \cite{DBLP:journals/corr/VinyalsL15,sordoni-EtAl:2015:NAACL-HLT,NIPS2015_MemN2N} in recent years.
These models are directly trained on past dialogs, without assumptions on the domain or dialog state structure \cite{bordes2016learning}.
One of their limitations is that they select responses only according to the content of the conversation and are thus incapable of adapting to users with different personalities.
Specifically, common issues with such content-based models include:
(i) the inability to \textbf{adjust language style} flexibly \cite{herzig-EtAl:2017:INLG2017};
(ii) the lack of a \textbf{dynamic conversation policy} based on the interlocutor's profile \cite{DBLP:journals/corr/JoshiMF17};
and (iii) the incapability of \textbf{handling ambiguities} in user requests. 

\begin{figure*}[t]
    \centering
    \begin{subfigure}[t]{\textwidth}
        \centering
        \includegraphics[width=\textwidth]{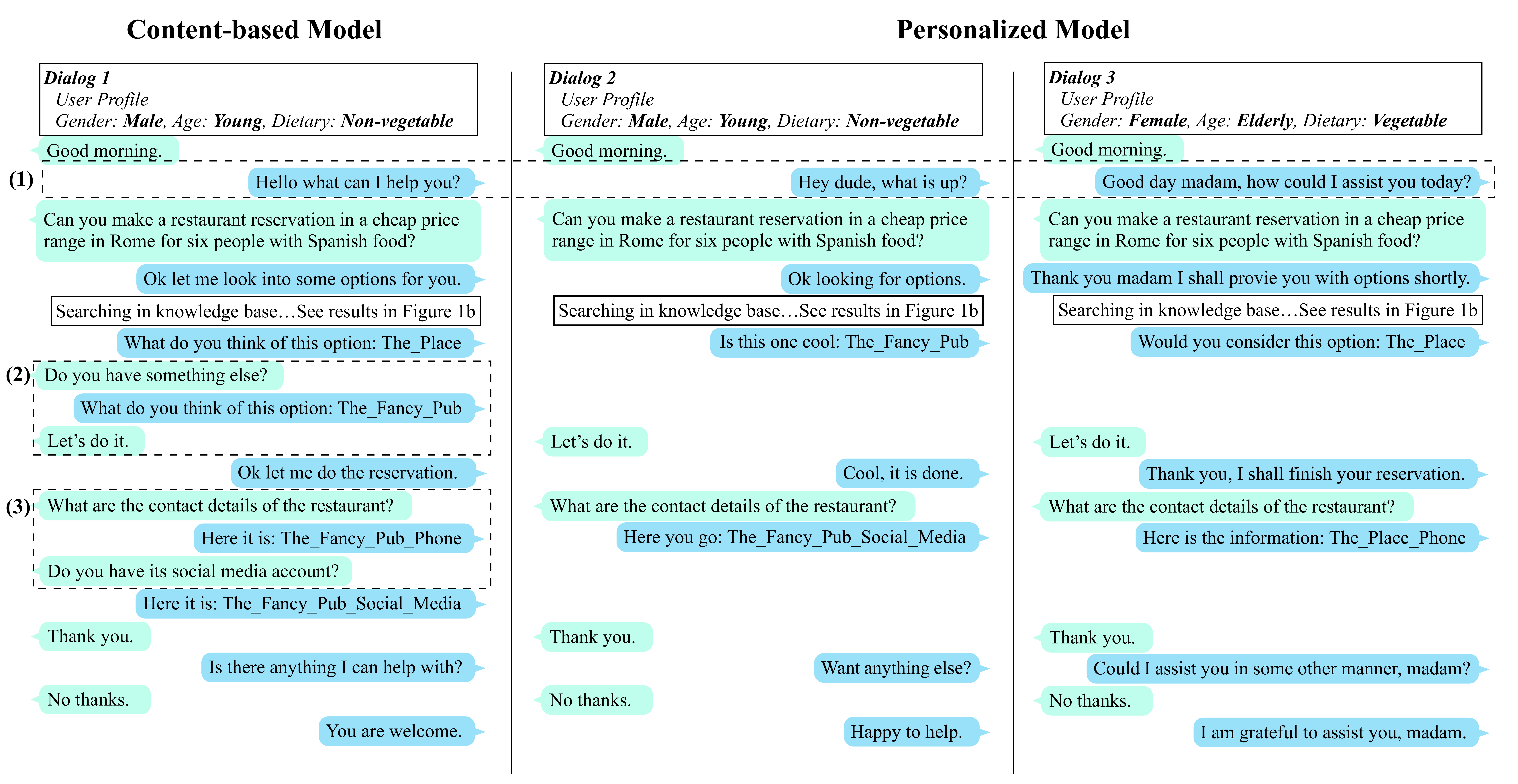}
        \caption{Example dialogs}
        \label{intro:example-dialogs}
    \end{subfigure}
    \begin{subfigure}[t]{\textwidth}
        \centering
        \includegraphics[width=0.8\textwidth]{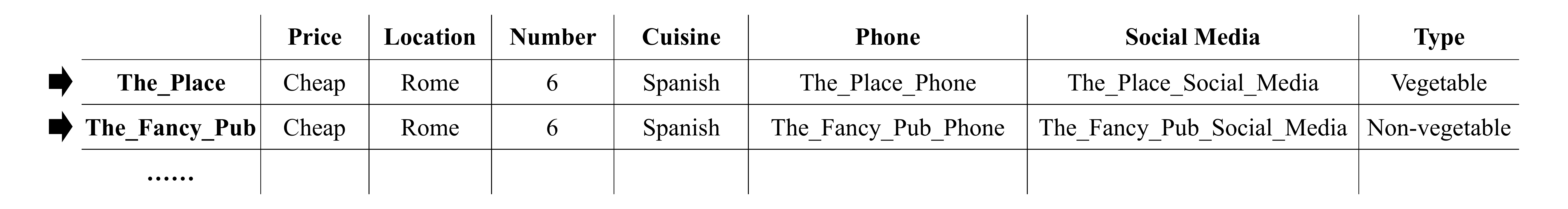}
        \caption{Searched results}
        \label{intro:example-kb}
    \end{subfigure}
    \caption{
        Examples to show the common issues with content-based models.
        We can see that the content-based model (1) is incapable of adjusting appellations and language styles, (2) fails to provide the best candidate, and (3) fails to choose the correct answer when facing ambiguities.
        (a) Three dialogs are chosen from the personalized bAbI dialog dataset.
        Personalized and content-based responses are generated by the \textsc{Personalized MemN2N} and a standard memory network, respectively.
        (b) Examples of valid candidates from a knowledge base that match the user request.
    }
    \label{intro:example}
\end{figure*}

Figure~\ref{intro:example} illustrates these problems with an example.
The conversation happens in a restaurant reservation scenario.
First, the responses from the content-based model are plain and boring, and not able to adjust appellations and language styles like the personalized model.
Second, in the recommendation phase, the content-based model can only provide candidates in a random order, while a personalized model can change recommendation policy dynamically, and in this case, match the user dietary.
Third, the word ``\textit{contact}'' can be interpreted into ``\textit{phone}'' or ``\textit{social media}'' contact information in the knowledge base.
Instead of choosing one randomly, the personalized model handles this ambiguity based on the learned fact that young people prefer social media account while the elders prefer phone number.

Psychologists have proven that during a dialog humans tend to adapt to their interlocutor to facilitate understanding, which enhances conversational efficiency \cite{brown1965social,brown1987theory,kroger1992rules}.
To improve agent intelligence, we may polish our model to learn such human behaviors in conversations.
A big challenge in building personalized dialog systems is how to utilize the user profile and generate personalized responses correspondingly.
To overcome it, existing works \cite{Qian2017AssigningPT,herzig-EtAl:2017:INLG2017} often conduct extra procedures to incorporate personalization in training, such as intermediate supervision and pre-training of user profiles, which are complex and time-consuming.
In contrast, our work is totally end-to-end.
    
    
In this paper, we propose a \textbf{\textsc{Profile Model}} and a \textbf{\textsc{Preference Model}} to leverage user profiles and preferences.
The \textsc{Profile Model} learns user personalities with distributed profile representation, and uses a global memory to store conversation context from other users with similar profiles.
In this way, it can choose a proper language style and change recommendation policy based on the user profile.
To address the problem of ambiguity, the \textsc{Preference Model} learns user preferences among ambiguous candidates by building a connection between the user profile and the knowledge base.
Since these two models are both under the \textsc{MemN2N} framework and make contributions to personalization in different aspects, we combine them into the \textbf{\textsc{Personalized MemN2N}}. 
    
    
Our experiments on a goal-oriented dialog corpus, the personalized bAbI dialog dataset, show that leveraging personal information can significantly improve the performance of dialog systems.
The \textsc{Personalized MemN2N} outperforms current state-of-the-art methods with over 7\% improvement in terms of per-response accuracy.
A test with real human users also illustrates that the proposed model leads to better outcomes, including higher task completion rate and user satisfaction.


    

\section{Related Work}
    
End-to-end neural approaches to building dialog systems have attracted increasing research interest.
It is well accepted that conversation agents include goal-oriented dialog systems and non goal-oriented (\textit{chit-chat}) bots.
    
Generative recurrent models like \textsc{Seq2Seq} have showed promising performance in non goal-oriented \textit{chit-chat} \cite{ritter-cherry-dolan:2011:EMNLP,DBLP:journals/corr/LowePSP15,Luo2018AnGeneration}.
More recently, retrieval-based models using a memory network framework have shown their potential in goal-oriented systems \cite{NIPS2015_MemN2N,bordes2016learning}.
Although steady progress has been made, there are still issues to be addressed:
most existing models are content-based, which are not aware of the interlocutor profile, and thus are not capable of adapting to different kinds of users.
Considerable research efforts have been devoted so far to make conversational agents smarter by incorporating user profile.
    
\textbf{Personalized Chit-Chat} 
The first attempt to model persona is \citet{li-EtAl:2016:P16-13}, which proposes an approach to assign specific personality and conversation style to agents based on learned persona embeddings.
\citet{DBLP:journals/corr/abs-1710-07388} describe an interesting approach that uses multi-task learning with personalized text data.
There are some researchers attempting to introduce personalized information to dialogs by transfer learning \cite{Yang:2017:PRG:3077136.3080706,DBLP:journals/corr/ZhangLWZ17}.
    
Since there is usually no explicit personalized information in conversation context, existing models \cite{Qian2017AssigningPT,herzig-EtAl:2017:INLG2017} often require extra procedures to incorporate personalization in training. 
\citet{Qian2017AssigningPT} add intermediate supervision to learn when to employ the user profile.
\citet{herzig-EtAl:2017:INLG2017} pre-train the user profile with external service.
This work, in contrast, is totally end-to-end.
    
A common approach to leveraging personality in these works is using a conditional language model as the response decoder \cite{ficler-goldberg:2017:StyVa,li-EtAl:2016:P16-13}.
This can help assign personality or language style to chit-chat bots, but it is useless in goal-oriented dialog systems.
Instead of assigning personality to agents \cite{li-EtAl:2016:P16-13,DBLP:journals/corr/abs-1710-07388,Qian2017AssigningPT}, our model pays more attention to the user persona and aims to make agents more adaptive to different kinds of interlocutors.
    
\textbf{Personalized Goal-Oriented Dialog} 
As most previous works \cite{li-EtAl:2016:P16-13,Liu2018ContentOrientedUM,Qian2017AssigningPT} focus on chit-chat, the combination of personalization and goal-oriented dialog remains unexplored.
Recently a new dataset has been released that enriches research resources for personalization in chit-chat \cite{Zhang2018PersonalizingDA}.
However, no open dataset allows researchers to train goal-oriented dialog with personalized information, until the personalized bAbI dialog corpus released by \citet{DBLP:journals/corr/JoshiMF17}.
    
Our work is in the vein of the memory network models for goal-oriented dialog from \citet{NIPS2015_MemN2N} and \citet{bordes2016learning}.
We enrich these models by incorporating the profile vector and using conversation context from users with similar attributes as global memory.
    
\section{End-to-End Memory Network}
\label{sec:MemN2N}
    
Since we construct our model based on the \textsc{MemN2N} by \citet{bordes2016learning}, we first briefly recall its structure to facilitate the delivery of our models.
    
The \textsc{MemN2N} consists of two components: context memory and next response prediction.
As the model conducts a conversation with the user, utterance (from the user) and response (from the model) are in turn appended to the memory.
At any given time step $t$ there are $\bm{c}_1^u, \cdots \bm{c}_t^u$ user utterances and $\bm{c}_1^r, \cdots \bm{c}_{t-1}^r$ model responses.
The aim at time $t$ is to retrieve the next response $\bm{c}_t^r$.
    
\textbf{Memory Representation}
Following \citet{DBLP:journals/corr/DodgeGZBCMSW15}, we represent each utterance as a bag-of-words using the embedding matrix $\bm{A}$, and the context memory $\bm{m}$ is represented as a vector of utterances as:
\begin{equation}
\begin{aligned}
\bm{m} = (\bm{A}\Phi(\bm{c}_1^u), \bm{A}\Phi(\bm{c}_1^r), \bm{A}\Phi(\bm{c}_2^u), \bm{A}\Phi(\bm{c}_2^r), \\
\cdots, \bm{A}\Phi(\bm{c}_{t-1}^u), \bm{A}\Phi(\bm{c}_{t-1}^r))
\end{aligned}
\end{equation}
where $\Phi(\cdot)$ maps the utterance to a bag of dimension $V$ (the vocabulary size), and $\bm{A}$ is a $d \times V$ matrix in which $d$ is the embedding dimension.
    
So far, information of which speaker spoke an utterance, and at what time during the conversation, are not included in the contents of memory.
We therefore encode those pieces of information in the mapping $\Phi$ by extending the vocabulary to contain $T = 1000$ extra ``time features'' which encode the index $i$ of an utterance into the bag-of-words, and two more features (\#$u$, \#$r$) encoding whether the speaker is the user or the bot.

The last user utterance $\bm{c}_t^u$ is encoded into $\bm{q} = \bm{A}\Phi(\bm{c}_t^u)$, which also denotes the initial query at time $t$, using the same matrix $\bm{A}$.
    
\textbf{Memory Operation}
The model first reads the memory to find relevant parts of the previous conversation for responses selection.
The match between $\bm{q}$ and the memory slots is computed by taking the inner product followed by a softmax: $\bm{\alpha}_i = \mathrm{Softmax}(\bm{q}^\top \bm{m}_i)$, which yields a vector of attention weights.
Subsequently, the output vector is constructed by $\bm{o} = \bm{R}\sum_i \bm{\alpha}_i \bm{m}_i$ where $\bm{R}$ is a $d \times d$ square matrix.
In a multi-layer \textsc{MemN2N} framework, the query is then updated with $\bm{q}_2 = \bm{q} + \bm{o}$.
Therefore, the memory can be iteratively reread to look for additional pertinent information using the updated query $\bm{q}_2$ instead of $\bm{q}$, and in general using $\bm{q}_k$ on iteration $k$, with a fixed number of iterations $N$ (termed $N$ hops).
    
Let $\bm{r}_i = \bm{W}\Phi(\bm{y}_i)$, where $\bm{W} \in \R^{d \times V}$ is another word embedding matrix, and $\bm{y}$ is a (large) set of candidate responses which includes all possible bot utterances and API calls.
The final predicted response distribution is then defined as:
\begin{equation}
\bm{\hat{r}} = \mathrm{Softmax}({\bm{q}_{N+1}}^\top \bm{r}_1, \cdots,{\bm{q}_{N+1}}^\top \bm{r}_C)
\label{eq:candidate_softmax}
\end{equation}
where there are $C$ candidate responses in $\bm{y}$.

\section{Personalized Dialog System}
\label{sec:model}

We first propose two personalized models.
The \textsc{Profile Model} introduces the personality of the interlocutor explicitly (using profile embedding) and implicitly (using global memory).
The \textsc{Preference Model} models user preferences over knowledge base entities.

The two models are independent to each other and we also explore their combination as the \textsc{Personalized MemN2N}.
Figure~\ref{PMemN2N:model} shows the structure of combined model. 
The different components are labeled with dashed boxes separately.
    
\subsection{Notation}

The user profile representation is defined as follows.
Each interlocutor has a user profile represented by $n$ attributes $\{ (k_i, v_i) \}_{i=1}^n$, where $k_i$ and $v_i$ denote the key and value of the $i$-th attribute, respectively.
Take the user in the first dialog in Figure~\ref{intro:example} as an example, the representation should be $\{ (\text{Gender}, \text{Male}), (\text{Age}, \text{Young}), (\text{Dietary}, \text{Non-vegetable}) \}$.
The $i$-th profile attribute is represented as a one-hot vector $\bm{a}_i \in \R^{d_i}$, where there are $d_i$ possible values for key $k_i$.
We define the user profile $\bm{\hat{a}} \in \R^{d^{(p)}}$ as the concatenation of one-hot representations of attributes: $\bm{\hat{a}} = \mathrm{Concat}(\bm{a}_1, \cdots , \bm{a}_n)$, where $d^{(p)} = \sum_i^n d_i$.
The notations of the memory network are the same as introduced in Section~\ref{sec:MemN2N}.
    
\subsection{Profile Model}
\label{sec:profile-model}

\begin{figure*}[t]
    \centering
    \includegraphics[width=\textwidth]{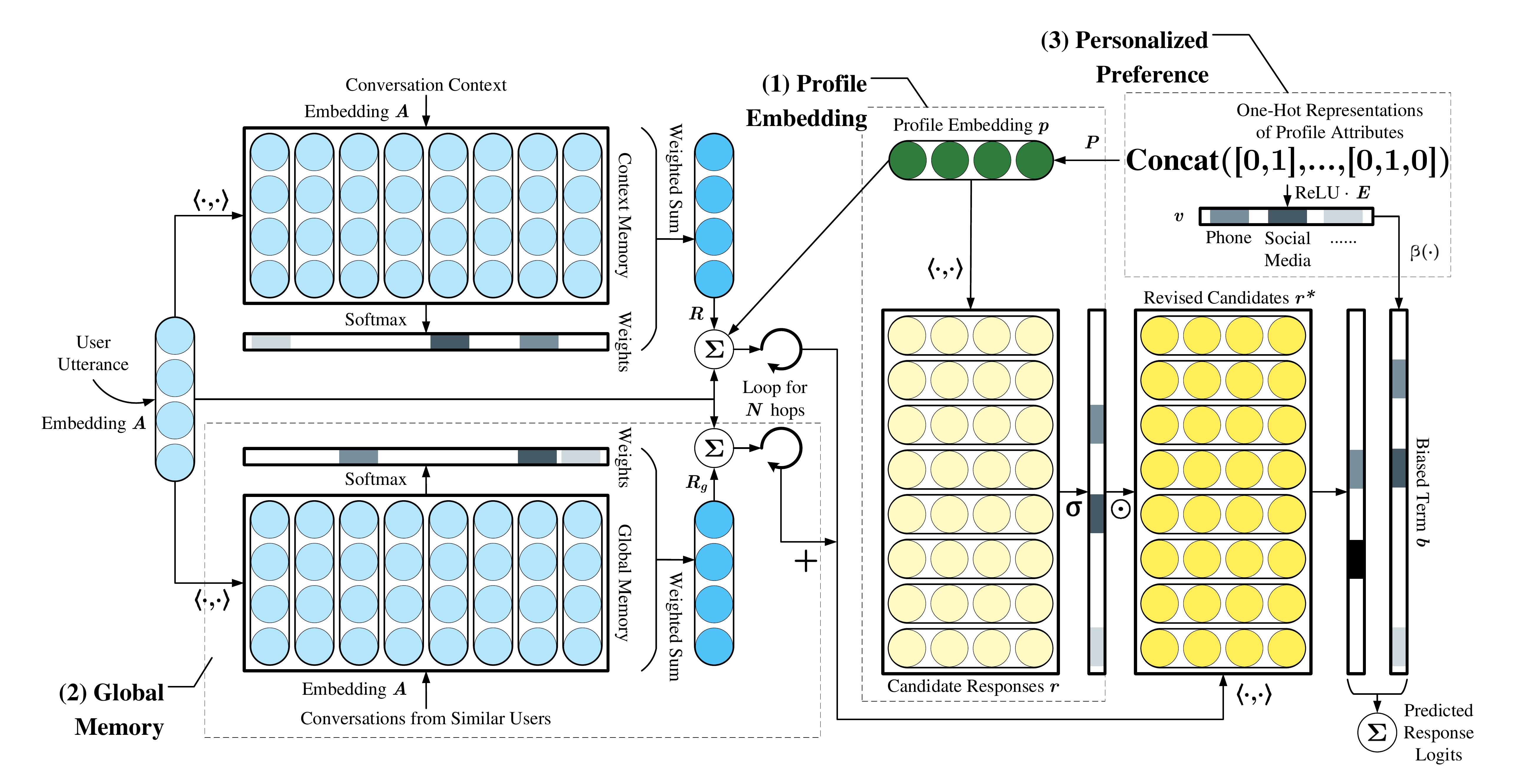}
    \caption{
        \textsc{Personalized MemN2N} architecture.
        The incoming user utterance is embedded into a query vector.
        The model first reads the memory (at top-left) to find relevant history and produce attention weights.
        Then it generates an output vector by taking the weighted sum followed by a linear transformation.
        Part (1) is \textbf{Profile Embedding}: the profile vector $\bm{p}$ is added to the query at each iteration, and is also used to revise the candidate responses $\bm{r}$.
        Part (2) is \textbf{Global Memory}: this component (at bottom-left) has an identical structure as the original \textsc{MemN2N}, but it contains history utterances from other similar users.
        Part (3) is \textbf{Personalized Preference}: the bias term is obtained based on the user preference and added to the prediction logits.
    }
    \label{PMemN2N:model}
\end{figure*}
    
Our first model is the \textsc{Profile Model}, which aims to integrate personalized information into the query and ranking part of the \textsc{MemN2N}.
The model consists of two different components: profile embedding and global memory.
    
\textbf{Profile Embedding} 
In the \textsc{MemN2N}, the query $\bm{q}$ plays a key role in both reading memory and choosing the response, while it contains no information about the user.
We expect to add a personalized information term to $\bm{q}$ at each iteration of the query.
Then, the model can be aware of the user profile in the steps of searching relevant utterances in the memory and selecting the final response from the candidates.
We thus obtain a distributed profile representation $\bm{p} \in \R^d$ by applying a linear transformation with the one-hot user profile: $\bm{p} = \bm{P} \bm{\hat{a}}$, where $\bm{P} \in \R^{d \times d^{(p)}}$.
Note that this distributed profile representation shares the same embedding dimension $d$ with the bag-of-words.
The query update equation can be changed as:
\begin{equation}
\bm{q}_{i+1} = \bm{q}_i + \bm{o}_i + \bm{p},
\end{equation}
where $\bm{q}_i$ and $\bm{o}_i$ are the query and output at the $i$-th hop, respectively.

Also, the likelihood of a candidate being selected should be affected directly by the user profile, no matter what the query is.
Therefore, we obtain tendency weights by computing the inner product between $\bm{p}$ and candidates followed by a sigmoid, and revise the candidates accordingly:
\begin{equation}
\label{eq:candidate_revised}
\bm{r}_i^* = \sigma( \bm{p}^\top \bm{r}_i ) \cdot \bm{r}_i,
\end{equation}
where $\sigma$ is a sigmoid. The prediction $\bm{\hat{r}}$ is then computed by Equation~\eqref{eq:candidate_softmax} using $\bm{r}^*$ instead of $\bm{r}$.
    
\textbf{Global Memory}
Users with similar profiles may expect the same or a similar response for a certain request.
Therefore, instead of using the profile directly, we also implicitly integrate personalized information of an interlocutor by utilizing the conversation history from similar users as a global memory.
The definition of similarity varies with task domains.
In this paper, we regard those with the same profile as similar users.
    
As shown in Figure~\ref{PMemN2N:model}, the global memory component has an identical structure as the original \textsc{MemN2N}.
The difference is that the contents in the memory are history utterances from other similar users, instead of the current conversation.
Similarly, we construct the attention weights, output vector, and iteration equation by
\begin{align}
\bm{\alpha}_i^{(g)} &= \mathrm{Softmax}(\bm{q}^\top \bm{m}_i^{(g)}) \\
\bm{o}^{(g)} &= \bm{R_g} \sum_i \bm{\alpha}_i^{(g)} \bm{m}_i^{(g)} \\
\bm{q}_{i+1}^{(g)} &= \bm{q}_i^{(g)} + \bm{o}_i^{(g)},
\end{align}
where $\bm{m}^{(g)}$ denotes the global memory, $\bm{\alpha}^{(g)}$ is the attention weight over the global memory, $\bm{R_g}$ is a $d \times d$ square matrix, $\bm{o}^{(g)}$ is the intermediate output vector and $\bm{q}_{i+1}^{(g)}$ is the result at the $i$-th iteration.
Lastly, we use $\bm{q}^+ = \bm{q} + \bm{q}^{(g)}$ instead of $\bm{q}$ to make the following computation.
    
\subsection{Preference Model}
\label{sec:preference-model}

The \textsc{Profile Model} has not yet solved the challenge of handling the ambiguity among KB entities, such as the choice between ``\textit{phone}'' and ``\textit{social media}'' in Figure~\ref{intro:example}.
The ambiguity refers to the user preference when more than one valid entities are available for a specific request.
We propose inferring such preference by taking the relation between user profile and knowledge base into account.

Assuming we have a knowledge base that describes the details of several items, where each row denotes an item and each column denotes one of their corresponding properties.
The entity $e_{i,j}$ at row $i$ and column $j$ is the value of the $j$-th property of item $i$.

The \textsc{Preference Model} operates as follows.
    
Given a user profile and a knowledge base with $K$ columns, we predict the user's preference on different columns.
We first model the user preference $\bm{v} \in \R^K$ as: 
\begin{equation}
\bm{v} = \mathrm{ReLU}(\bm{E} \bm{\hat{a}})
\end{equation}
where $\bm{E} \in \R^{K \times d^{(p)}}$.
Note that we assume the bot cannot provide more than one option in a single response, so a candidate can only contains one entity at most.
The probability of choosing a candidate response should be affected by this preference if the response mentions one of the KB entities.

We add a bias term $\bm{b} = \beta(\bm{v}, \bm{r}, \bm{m}) \in \R^C$ to revise the logits in Equation~\eqref{eq:candidate_softmax}.
The bias for $k$-th candidate $\bm{b}_k$ is constructed as the following steps. If the $k$-th candidate contains no entity, then $\bm{b}_k = 0$; if the candidate contains an entity $e_{i,j}$, which belongs to item $i$, then $\bm{b}_k = \lambda(i, j)$, where given the current conversation context $ctx$,
\begin{equation}
\lambda(i,j) = \begin{cases}
\bm{v}_j, & \text{if item } i \text{ is mentioned in } ctx; \\
0, & \text{otherwise}.
\end{cases}
\end{equation}
For example, the candidate ``\textit{Here is the information: The\_Place\_Phone}'' contains a KB entity ``\textit{The\_Place\_Phone}'' which belongs to restaurant ``\textit{The\_Place}'' and column ``\textit{Phone}''.
If ``\textit{The\_Place}'' has been mentioned in the conversation, the bias term for this response should be $\bm{v}_{Phone}$.
    
We update the Equation~\eqref{eq:candidate_softmax} to
\begin{equation}
\bm{\hat{r}} = \mathrm{Softmax}({\bm{q}_{N+1}}^\top \bm{r}_1 + \bm{b}_1, \cdots, {\bm{q}_{N+1}}^\top \bm{r}_C + \bm{b}_C).
\label{eq:biased_candidate_softmax}
\end{equation}
    
\subsection{Combined Model}

As discussed previously, the \textsc{Profile Model} and the \textsc{Preference Model} make contributions to personalization in different aspects.
The \textsc{Profile Model} enables the \textsc{MemN2N} to change the response policy based on the user profile, but fails to establish a clear connection between the user and the knowledge base.
On the other hand, the \textsc{Preference Model} bridges this gap by learning the user preferences over the KB entities.

To take advantages of both models, we construct a general \textsc{Personalized MemN2N} model by combining them together, as shown in Algorithm~\ref{algo}.
All these models are trained to minimize a standard cross-entropy loss between $\bm{\hat{r}}$ and the true label $\bm{r}_{true}$.
    
\begin{algorithm}
    \caption{Response Prediction by \textsc{Personalized MemN2N}}
    \label{algo}
    
    \textbf{Input:} User utterance $\bm{q}$, Context memory $\bm{m}$, global memory $\bm{m}^{(g)}$, candidates $\bm{r}$ and user profile $\bm{\hat{a}}$
    
    \textbf{Output:} The index $y$ of the next response
    \begin{algorithmic}[1]
        \Procedure{Predict}{$\bm{q},\bm{m},\bm{m}^{(g)},\bm{r},\bm{\hat{a}}$}
        \State $\bm{p} \gets \bm{P} \bm{\hat{a}}$
        \Comment{Profile embedding}
        \State $\bm{q}^{(g)} \gets \bm{q}$
        
        \For{$N$ hops}
        \State $\bm{\alpha} \gets$ $\mathrm{Softmax}(\bm{q}^\top \bm{m})$
        \State $\bm{q} \gets \bm{q} + \bm{p} + \bm{R} \sum_i \bm{\alpha}_i \bm{m}_i$
        
        \State $\bm{\alpha}^{(g)} \gets$ $\mathrm{Softmax}((\bm{q}^{(g)})^\top \bm{m}^{(g)})$
        \State $\bm{q}^{(g)} \gets \bm{q}^{(g)} + \bm{R_g} \sum_i \bm{\alpha}_i^{(g)} \bm{m}_i^{(g)}$
        \EndFor
        
        \State $\bm{v} = \mathrm{ReLU}(\bm{E} \bm{\hat{a}})$
        \State $\bm{b} = \beta(\bm{v}, \bm{r}, \bm{m})$
        \Comment{Bias term}
        \State $\bm{q}^+ = \bm{q} + \bm{q}^{(g)}$
        \Comment{Final query}
        \State $\bm{r}^* = \sigma(\bm{p}^\top \bm{r}) \odot \bm{r}$
        \Comment{Revised candidates}
        \State $\bm{\hat{r}}_i \gets \mathrm{Softmax}((\bm{q}^+)^\top \bm{r}^*_i + \bm{b}_i)$
        \State $y \gets \argmax_i \bm{\hat{r}}_i$
        
        \EndProcedure
    \end{algorithmic}
\end{algorithm}
    
\section{Experiments}
    
\subsection{Dataset}
    
The personalized bAbI dialog dataset \cite{DBLP:journals/corr/JoshiMF17} is a multi-turn dialog corpus extended from the bAbI dialog dataset \cite{bordes2016learning}. 
It introduces an additional user profile associated with each dialog and updates the utterances and KB entities to integrate personalized style.
Five separate tasks in a restaurant reservation scenario are introduced along with the dataset.
Here we briefly introduce them for better understanding of our experiments.
More details on the dataset can be found in the work by \citet{DBLP:journals/corr/JoshiMF17}.
    
\textbf{Task~1: Issuing API Calls}
Users make queries that contain several blanks to fill in. 
The bot must ask proper questions to fill the missing fields and make the correct API calls.
    
\textbf{Task~2: Updating API Calls}
Users may update their request and the bot must change the API call accordingly.
    
\textbf{Task~3: Displaying Options}
Given a user request, the KB is queried and the returning facts are added to the dialog history.
The bot is supposed to sort the options based on how much users like the restaurant.
The bot must be conscious of the user profile and change the sorting strategy accordingly to accomplish this task.
    
\textbf{Task~4: Providing Information}
Users ask for some information about a restaurant, and more than one answer may meet the requirement (i.e., contact with-respect-to social media account and phone number).
The bot must infer which answer the user prefers based on the user profile.
    
\textbf{Task~5: Full Dialog}
This task conducts full dialog combining all the aspects of Tasks~1 to 4.
    

The difficulties of personalization in these tasks are not incremental. 
In Tasks~1 and 2, the bot is only required to select responses with appropriate meaning and language style. 
In Tasks~3 and 4, the knowledge base is supposed to be searched, which makes personalization harder.
In these two tasks, apart from capturing shallow personalized features in the utterances such as language style, the bot also has to learn different searching or sorting strategies for different user profiles. 
In Task~5 we expect an average performance (utterance-wise) since it combines the other four tasks.

\begin{table*}[t]
    \centering
    \small
    \begin{tabular}{lccccc}
        \toprule
        & \textbf{T1: Issuing} & \textbf{T2: Updating} & \textbf{T3: Displaying} & \textbf{T4: Providing} & \textbf{T5: Full} \\
        \textbf{Models} & \textbf{API Calls} & \textbf{API Calls} & \textbf{Options} & \textbf{Information} & \textbf{Dialog} \\
        \midrule
        1. Supervised Embeddings & 84.37  & 12.07  & 9.21  & 4.76  & 51.60  \\
        2. MemN2N & 99.83 (98.87) & \textbf{99.99} (99.93) & 58.94 (58.71) & 57.17 (57.17) & 85.10 (77.74) \\
        3. Split MemN2N & 85.66 (82.44) & 93.42 (91.27) & 68.60 (68.56) & 57.17 (57.11) & 87.28 (78.10) \\
        \midrule
        4. Profile Embedding & \textbf{99.96 (99.98)} & 99.96 (99.94) & 71.00 (70.95) & 57.18 (57.18) & 93.83 (81.32) \\
        5. Global Memory & 99.76 (98.96) & 99.93 (99.74) & 71.01 (71.11) & 57.18 (57.18) & 91.70 (81.43) \\
        6. Profile Model & 99.93 (99.96) & 99.94 (99.94) & 71.12 (70.78) & 57.18 (57.18) & 93.91 (82.57) \\
        \midrule
        7. Preference Model & 99.80 (99.95) & 99.97 (\textbf{99.97}) & 68.90 (68.34) & 81.38 (80.30) & 94.97 (86.56) \\
        \midrule
        8. Personalized MemN2N & 99.91 (99.93) & 99.94 (99.95) & \textbf{71.43 (71.52)} & \textbf{81.56 (80.79)} & \textbf{95.33 (88.07)} \\
        \bottomrule
    \end{tabular}%
    \caption{
        Evaluation results of the \textsc{Presonalized MemN2N} on the personalized bAbI dialog dataset.
        Rows~1 to 3 are baseline models.
        Rows~4 to 6 are the \textsc{Profile Model} with profile embedding, global memory and both of them, respectively.
        In each cell, the first number represents the per-response accuracy on the full set, and the number in parenthesis represents the accuracy on a smaller set with 1000 dialogs.
    }
    \label{exper:results}
\end{table*}

There are two variations of dataset provided for each task: a full set with around $6000$ dialogs and a small set with only $1000$ dialogs to create realistic learning conditions.
We get the dataset released on ParlAI.\footnote{\url{http://parl.ai/}}
    
\subsection{Baselines}
    
We consider the following baselines:

\textbf{Supervised Embedding Model}:
a strong baseline for both chit-chat and goal-oriented dialog \cite{DBLP:journals/corr/DodgeGZBCMSW15,bordes2016learning}.
    
\textbf{Memory Network}:
the \textsc{MemN2N} by \citet{bordes2016learning}, which has been described in detail in Section~\ref{sec:MemN2N}.
We add the profile information as an utterance said by the user at the beginning of each dialog.
In this way the standard \textsc{MemN2N} may capture the user persona to some extent.
    
\textbf{Split Memory Network}:
the model proposed by \citet{DBLP:journals/corr/JoshiMF17} that splits the memory into two parts: profile attributes and conversation history.
The various attributes are stored as separate entries in the profile memory before the dialog starts, 
and the conversation memory operates the same as the \textsc{MemN2N}.
    
    
\subsection{Experiment Settings}

The parameters are updated by Nesterov accelerated gradient algorithm \cite{nesterov1983method} and initialized by Xavier initializer.
We try different combinations of hyperparameters and find the best settings as follows.
The learning rate is $0.001$, and the parameter of momentum $\gamma$ is $0.9$.
Gradients are clipped to avoid gradient explosion with a threshold of $10$.
We employ early-stopping as a regularization strategy.
Models are trained in mini-batches with a batch size of $64$.
The dimensionality of word/profile embeddings is $128$.
We set the maximum context memory and global memory size (i.e. number of utterances) as $250$ and $1000$, separately.
We pad zeros if the number of utterances in a memory is less than $250$ or $1000$, otherwise we keep the last $250$ utterances for the context memory, or randomly choose $1000$ valid utterances for the global memory.
    
\subsection{Results}

Following \citet{DBLP:journals/corr/JoshiMF17}, we report per-response accuracy across all models and tasks on the personalized bAbI dataset in Table~\ref{exper:results}. 
The per-response accuracy counts the percentage of correctly chosen candidates.
    
\subsubsection{Profile Model}
    

Rows~4 to 6 of Table~\ref{exper:results} show the evaluation results of the \textsc{Profile Model}.

As reported in \citet{DBLP:journals/corr/JoshiMF17}, their personalized dialogs model might be too complex for some simple tasks (such as Tasks~1 and 2, which do not rely on KB facts) and tends to overfit the training data.
It is reflected in the failure of the split memory model on Tasks~1 and 2.
Although it outperforms the standard \textsc{MemN2N} in some complicated tasks, the latter one is good enough to capture the profile information given in a simple raw text format, and defeats the split memory model in simpler tasks.
    
To overcome such a challenge, we avoid using excessively complex structures to model the personality.
Instead, we only represent the profile as an embedding vector or implicitly.
As expected, both profile embedding and global memory approach accomplish Tasks~1 and 2 with a very high accuracy and also notably outperform the baselines in Task~3, which requires utilizing KB facts along with the profile information.
Also, the performance of combining the two components together, as shown in row~6, is slightly better than using them independently.
The result suggests that we can take advantages of using profile information in an explicit and implicit way in the meantime.
    
    
\subsubsection{Preference Model}
Since the \textsc{Profile Model} does not build a clear connection between the user and the knowledge base, as discussed in Section~\ref{sec:model}, it may not solve ambiguities among the KB columns. 
The experiment results are consistent with this inference: the performance of the \textsc{Profile Model} on Task~4, which requires user request disambiguation, is particularly close to the baselines.
    
Row~7 shows the evaluation results of the \textsc{Preference Model}, which is proposed to handle the above mentioned challenge.
The model achieves significant improvements on Task~4 by introducing the bias term derived from the learned user preference.

Besides, the restaurant sorting challenge in Task~3 depends on the properties of a restaurant to some extent.
Intuitively, different properties of the restaurants are weighted differently, and the user preference over the KB columns can be considered as scoring weights which is useful for task-solving.
As a result, the model also improves the performance in Task~3 compared to the standard \textsc{MemN2N}.
    
    
\subsubsection{Personalized MemN2N}

We test the performance of the combined \textsc{Personalized MemN2N} as well.
As we have analyzed in Section~\ref{sec:model}, the \textsc{Profile Model} and the \textsc{Preference Model} make contributions to personalization in different aspects and their combination has the potential to take advantages of both models.
Experiment results confirm our hypothesis that the combined model achieves the best performance with over 7\% (and 9\% on small sets) improvement over the best baseline for the full dialog task (Task~5).
    
\section{Analysis}

As the proposed \textsc{Personalized MemN2N} achieves better performance than previous approaches, we conduct an analysis to gain further insight on how the integration of profile and preference helps the response retrieval. 

\subsection{Analysis of Profile Embeddings}

Since we use the learned profile embeddings to obtain tendency weights for candidates selection, as is illustrated in Equation~\eqref{eq:candidate_revised}, we expect to observe larger weights on candidates that correctly match the profile. 
For instance, given a profile ``\textit{Gender: Male, Age: Young}'', we can generate a weight for each response candidate.
Due to the fact that candidates are collected from dialogs with different users, they can be divided based on the user profile.
Those candidates in the group of \textit{young male} should have larger weights than others.

\begin{figure}[htb]
    \centering
    \includegraphics[height=3.5cm]{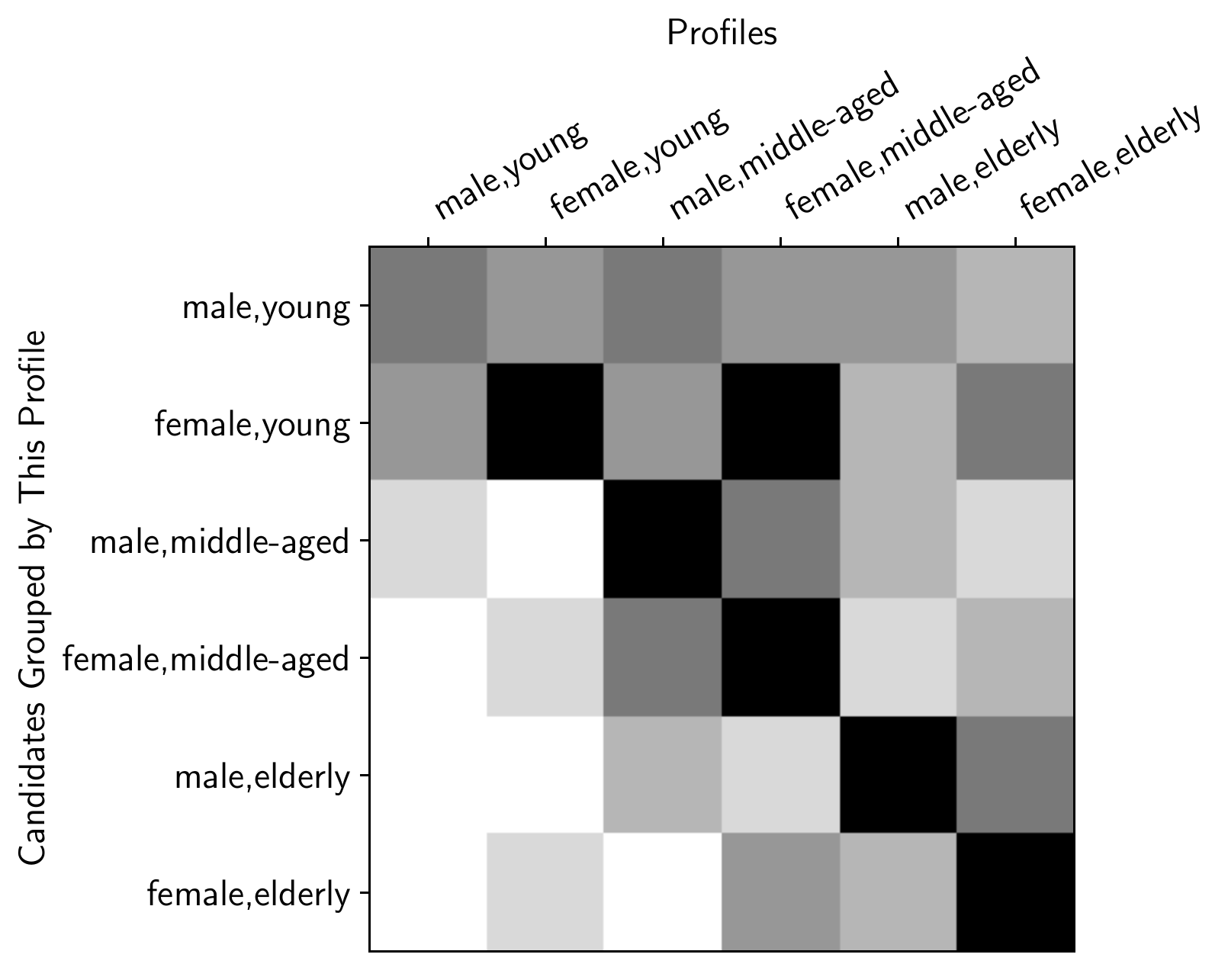}
    \caption{
        Confusion matrix for profiles and generated tendency weights.
        Darker cell means larger weight value.
    }
    \label{ana:confusion-matrix}
\end{figure}

We group the candidates by their corresponding user profile.
For each profile, we generate tendency weights and collect the average value for each group.
Figure~\ref{ana:confusion-matrix} visualizes the results by a confusion matrix.
The weights on the diagonal are significantly larger than others, which demonstrates the contribution of profile embeddings in candidate selection.

\subsection{Analysis of Global Memory}

\begin{table}[t]
    \centering
    \small
    \begin{tabular}{lc}
         \toprule
         \textbf{Models} & \textbf{T5: Full Dialog} \\
         \midrule
         Global Memory (similar users) & 91.70 (81.43) \\
         Global Memory (random users) & 87.17 (78.02) \\
         \bottomrule
    \end{tabular}
    \caption{
        Evaluation results of the control experiment on Task~5: Full Dialog.
    }
    \label{ana:global}
\end{table}

To better illustrate how much the global memory impacts the performance of the proposed model, we conduct a control experiment.
Specifically, we build a model with the same global memory component as described in Section~\ref{sec:profile-model}, but the utterances in the memory are from randomly chosen users rather than similar users.
We report the results of the control experiment on Task~5 in Table~\ref{ana:global}.
The numbers indicate that the global memory does help improve the performance.

\subsection{Analysis of Preference}

\begin{figure}[t]
    \centering
    \begin{subfigure}[b]{0.48\linewidth}
        \centering
        \includegraphics[width=0.95\linewidth]{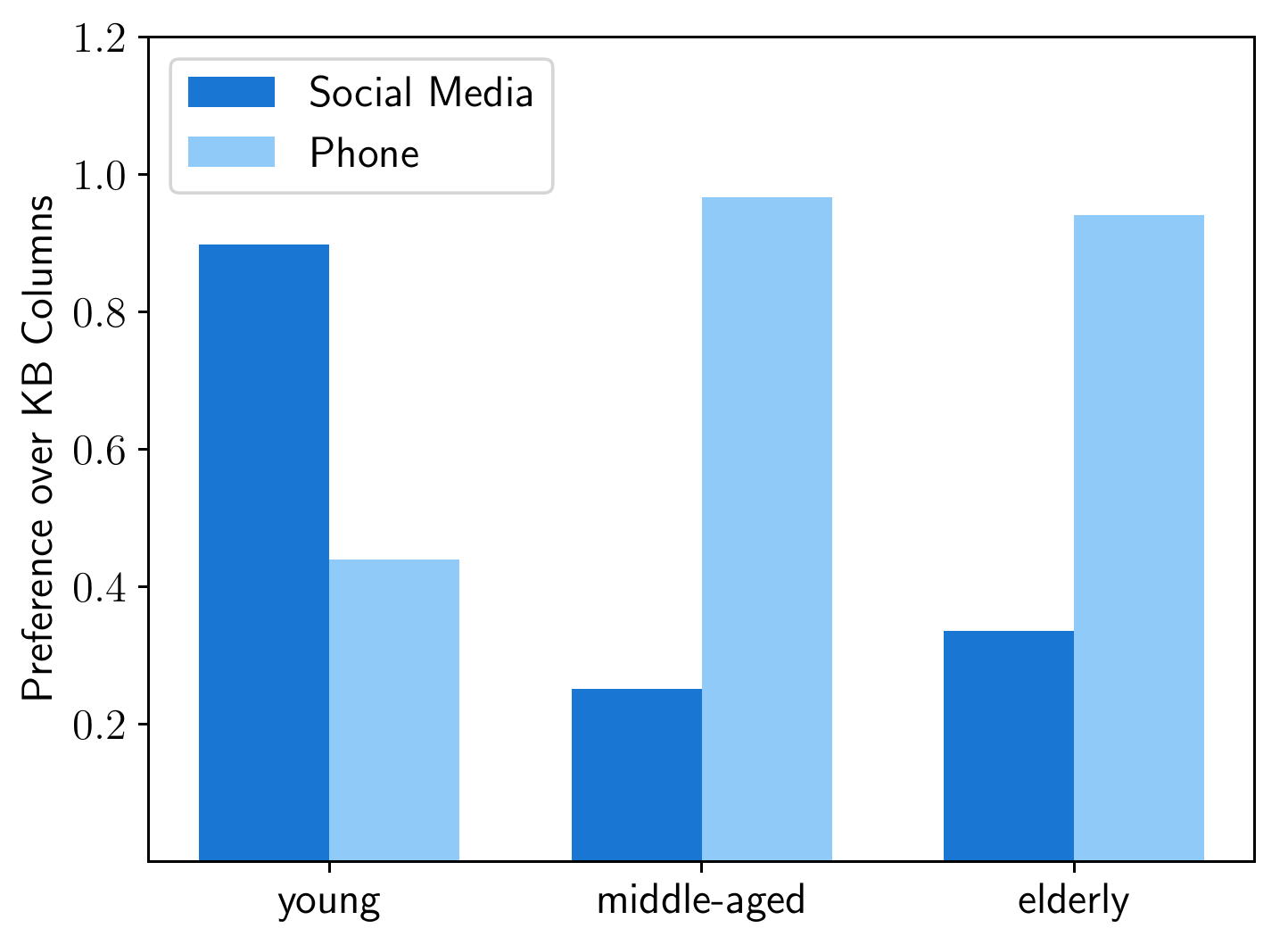}
        \caption{Task~4}
    \end{subfigure}
    \begin{subfigure}[b]{0.48\linewidth}
        \centering
        \includegraphics[width=0.95\linewidth]{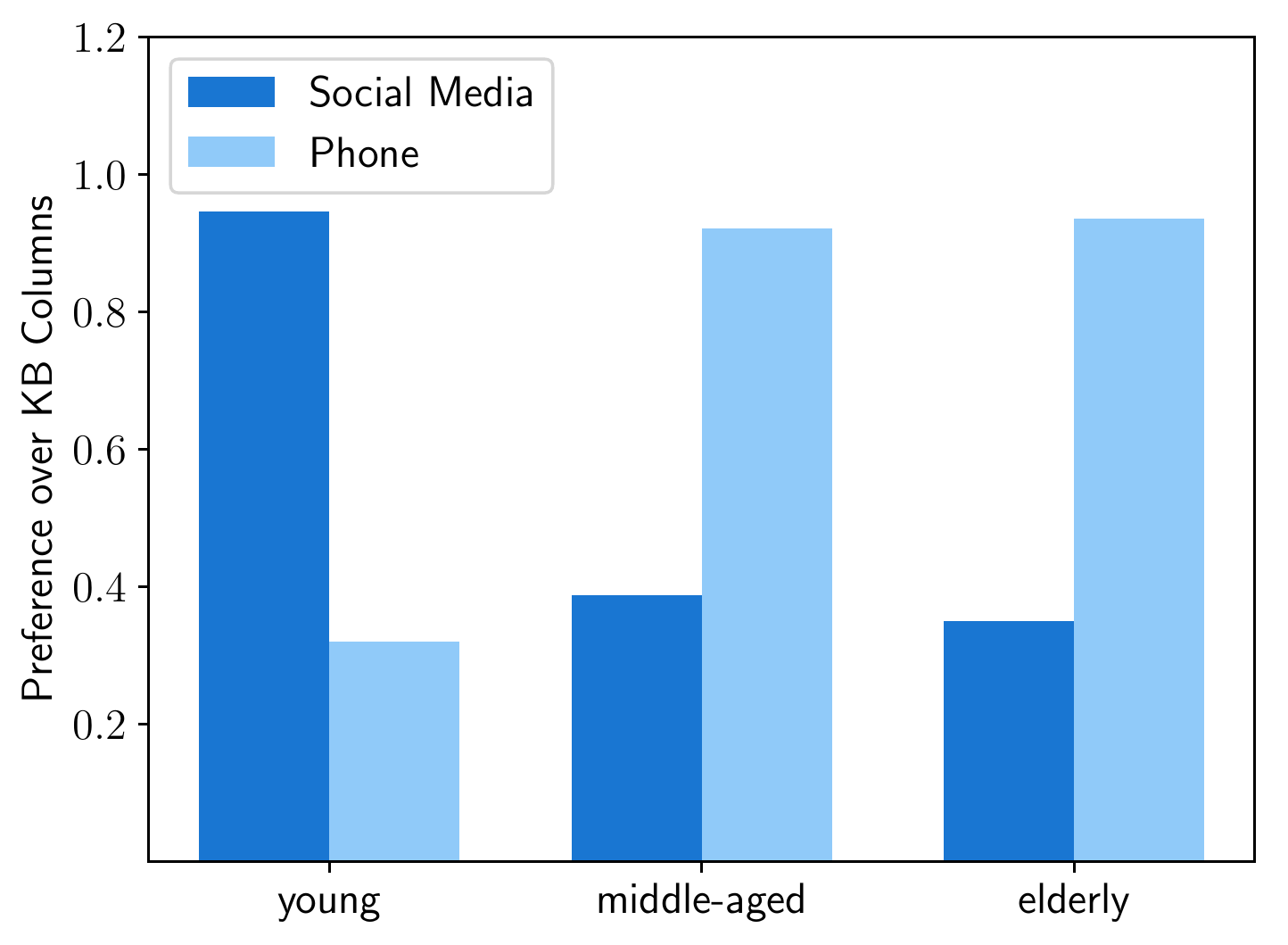}
        \caption{Task~5}
    \end{subfigure}
    \caption{
        The preference arguments learned from Task~4: Providing Information, and Task~5: Full Dialog.
        The preference score is computed by an L2 normalization from $\bm{v}$.
    }
    \label{exper:preference-visualization}
\end{figure}

Remember that we use a preference vector $\bm{v}$ to represent the user's preference over the columns in the knowledge base.
Therefore, we investigate the learned arguments grouped by profile attributes.
As seen in Figure~\ref{exper:preference-visualization}, the model successfully learns the fact that young people prefer social media as their contact information, while middle-aged and elderly people prefer phone number.
The result shows great potential and advantage of end-to-end models. They are capable of learning meaningful intermediate arguments while being much simpler than existing reinforcement learning methods and pipeline models for the task of personalization in dialogs.

\subsection{Human Evaluation}

To demonstrate the effectiveness of the personalization approach over standard models more convincingly, we build an interactive system based on the proposed model and baselines, and conduct a human evaluation.
Since it is impractical to find testers with all profiles we need, we randomly build $20$ profiles with different genders, ages and preferences, and ask three judges to act as the given roles.
They talk to the system and score the conversations in terms of task completion rate and satisfaction.
Task completion rate stands for how much the system accomplish the users' goal.
Satisfaction refers to whether the responses are appropriate to the user profile.
The scores are averaged and range from $0$ to $1$ ($0$ is the worst and $1$ is perfect).
We find that \textsc{Personalized MemN2N} wins the \textsc{MemN2N} baseline with $27.6\%$ and $14.3\%$ higher in terms of task completion rate and satisfaction, respectively, with $p < 0.03$.



\section{Conclusion and Future Work}

We introduce a novel end-to-end model for personalization in goal-oriented dialog.
Experiment results on open datasets and further analysis show that the model is capable of overcoming some existing issues in dialog systems. 
The model improves the effectiveness of the bot responses with personalized information, and thus greatly outperforms state-of-the-art methods.

In future work, more representations of personalities apart from the profile attribute can be introduced into goal-oriented dialogs models.
Besides, we may explore on learning profile representations for non-domain-specific tasks and consider KB with more complex format such as ontologies.

\section*{Acknowledgements}

We thank all reviewers for providing the constructive suggestions.
Also thanks to Danni Liu, Haoyan Liu and Yuanhao Xiong for the helpful discussion and proofreading.
Xu Sun is the corresponding author of this paper.

\bibliography{personalized}
\bibliographystyle{aaai}

\end{document}